%% file: main.tex
\documentclass[]{bytedance}
\usepackage[toc,page,header]{appendix}

\usepackage{minitoc}
\usepackage{amsfonts}
\usepackage{amssymb}
\usepackage{tabularx}
\usepackage{listings}
\usepackage{xcolor}
\usepackage{cancel}

\usepackage{tabulary,multirow,xspace}
\usepackage{fixmath,mathtools,nicefrac,mmstyle}
\usepackage{subcaption}
\usepackage{caption}
\usepackage{wrapfig} 
\usepackage[misc]{ifsym} 
\usepackage{colortbl}

\usepackage{wrapfig}
\usepackage{multicol}
\usepackage[most]{tcolorbox}
\usepackage{pifont}

\definecolor{codegreen}{rgb}{0,0.6,0}
\definecolor{codegray}{rgb}{0.5,0.5,0.5}
\definecolor{codepurple}{rgb}{0.58,0,0.82}
\definecolor{backcolour}{rgb}{0.95,0.95,0.92}
\definecolor{boxblue}{RGB}{57,89,163}
\definecolor{boxbluebg}{RGB}{230,237,250} 

\lstdefinestyle{mystyle}{
    backgroundcolor=\color{backcolour},   
    commentstyle=\color{codegreen},
    keywordstyle=\color{magenta},
    numberstyle=\tiny\color{codegray},
    stringstyle=\color{codepurple},
    basicstyle=\ttfamily\footnotesize,
    breakatwhitespace=false,         
    breaklines=true,                 
    captionpos=b,                    
    keepspaces=true,                 
    numbers=none,                    
    numbersep=5pt,                  
    showspaces=false,                
    showstringspaces=false,
    showtabs=false,                  
    tabsize=2
}
\definecolor{mygray1}{gray}{.95}
\definecolor{mygray2}{gray}{.9}
\definecolor{mygray3}{gray}{.95}
\usepackage{pifont}

\newlength\savewidth
\newcolumntype{x}[1]{>{\centering\arraybackslash}p{#1pt}}

\newcommand{\app}{\raise.17ex\hbox{$\scriptstyle\sim$}}

\usepackage{xcolor}
\input{macros}

\usepackage{listings}

\definecolor{carnelian}{rgb}{0.7, 0.11, 0.11}
\usepackage{booktabs}
\usepackage{makecell}
\usepackage{multirow}

\usepackage{adjustbox}
\usepackage{cuted} 

\usepackage{algorithm}
\usepackage{algorithmic}
\definecolor{myblue}{RGB}{210, 225, 255}
\definecolor{mytextblue}{RGB}{51, 161, 201}
\definecolor{mypurple}{RGB}{218, 112, 214}

\definecolor{commentgreen}{rgb}{0.1, 0.4, 0.1}
\definecolor{keywordblue}{rgb}{0.1, 0.1, 0.7}
\definecolor{stringred}{rgb}{0.7, 0.1, 0.1}

\lstdefinestyle{mystyle}{
    commentstyle=\color{commentgreen},
    keywordstyle=\color{keywordblue},   
    stringstyle=\color{stringred},
    basicstyle=\ttfamily\scriptsize, 
    breaklines=true,
    keepspaces=true,
    showstringspaces=false,
    frame=none,                     
    language=Python, 
}

\newcommand{\name}{OmniInsert}
\title{\name{}: Mask-Free Video Insertion of Any Reference via Diffusion Transformer Models}

\author{
\centerline{
Jinshu Chen $^*$ \quad 
Xinghui Li $^{*\dagger}$ \quad  
Xu Bai $^*$ \quad 
Tianxiang Ma \quad
Pengze Zhang \quad
} 
\centerline{
Zhuowei Chen \quad
Gen Li \quad
Lijie Liu \quad
Songtao Zhao $^{\dagger}$ \quad
Bingchuan Li $^{\ddagger}$ \quad
Qian He \quad
}
}

\affiliation[]{Intelligent Creation Lab, ByteDance}

\footnotetext{* Equal Contribution, $\dagger$ Corresponding author, $\ddagger$ Project lead.}

\input{sec/1_abs}

\date{\today}

\checkdata[Project Page]{\url{https://phantom-video.github.io/OmniInsert/}}

\begin{document}
\maketitle

\input{sec/2_intro}
\input{sec/3_related}
\input{sec/4_methods}

\input{sec/5_exps}
\input{sec/6_conclusion}

\clearpage

\bibliographystyle{plainnat}
\bibliography{main}

\clearpage
\input{supp_sections/0_implementation}    
\input{supp_sections/1_bench}
\input{supp_sections/2_more_res}
\input{supp_sections/3_limitations}

\end{document}

%% file: macros.tex
\usepackage{graphicx}
\usepackage{amssymb}
\usepackage{pifont}
\usepackage{floatrow}
\usepackage{amsmath} 
\usepackage{float}
\usepackage{wrapfig}
\usepackage{multirow}
\usepackage{tcolorbox}
\tcbuselibrary{breakable, skins, raster}
\usepackage{listings}

%% file: sec/1_abs.tex
\abstract{
       Recent advances in video insertion based on diffusion models are impressive. However, existing methods rely on complex control signals but struggle with subject consistency, limiting their practical applicability. In this paper, we focus on the task of \textit{\textbf{Mask-free Video Insertion}}, and aim to resolve three key challenges: data scarcity, subject–scene equilibrium, and insertion harmonization. To address the data scarcity, we propose a new data pipeline \textit{\textbf{InsertPipe}}, constructing diverse cross-pair data automatically. Building upon our data pipeline, we develop \textit{\textbf{OmniInsert}}, a novel unified framework for mask-free video insertion from both single and multiple subject references. Specifically, to maintain subject-scene equilibrium, we introduce a simple yet effective Condition-Specific Feature Injection mechanism to distinctly inject multi-source conditions and propose a novel Progressive Training strategy that enables the model to balance feature injection from subjects and source video. Meanwhile, we design the Subject-Focused Loss to improve the detailed appearance of the subjects. To further enhance insertion harmonization, we propose an Insertive Preference Optimization methodology to optimize the model by simulating human preferences, and incorporate a Context-Aware Rephraser module during reference to seamlessly integrate the subject into the original scenes. To address the lack of a benchmark for the field, we introduce \textit{\textbf{InsertBench}}, a comprehensive benchmark comprising diverse scenes with meticulously selected subjects. Evaluation on InsertBench indicates \textit{OmniInsert} outperforms state-of-the-art closed-source commercial solutions. Code will be released.
}

%% file: sec/2_intro.tex
\section{Introduction}
The emergence of diffusion models~\cite{ho2020denoising, peebles2023scalable} has significantly advanced the field of video generation. In recent years, numerous open-source and commercial video generation models have been developed~\cite{hong2022cogvideo, wan2025wan, gao2025seedance}, providing powerful tools for creating video content. Beyond generation, video editing naturally arises as an extension, focusing on  re-generate a reference video under fine-grained control signals like depth~\cite{esser2023structure, liao2020dvi}, pose~\cite{hu2024animate, ma2024follow, karras2023dreampose, xu2024magicanimate}, motion~\cite{jeong2023ground, feng2024ccedit, liang2024flowvid, liew2023magicedit} and point~\cite{teng2023drag, deng2024dragvideo}. Among various editing techniques, reference-image-based editing~\cite{ku2024anyv2v, jiang2025vace} plays a key role by providing explicit visual cues to guide the editing process. Specially, the Video Insertion (VI) task of inserting a reference subject into the source video has sparked considerable research interest, due to its significant potential for practical applications in film production, advertising, and artistic design.

\begin{figure}[H]
    \centering	
    \includegraphics[width=\textwidth]{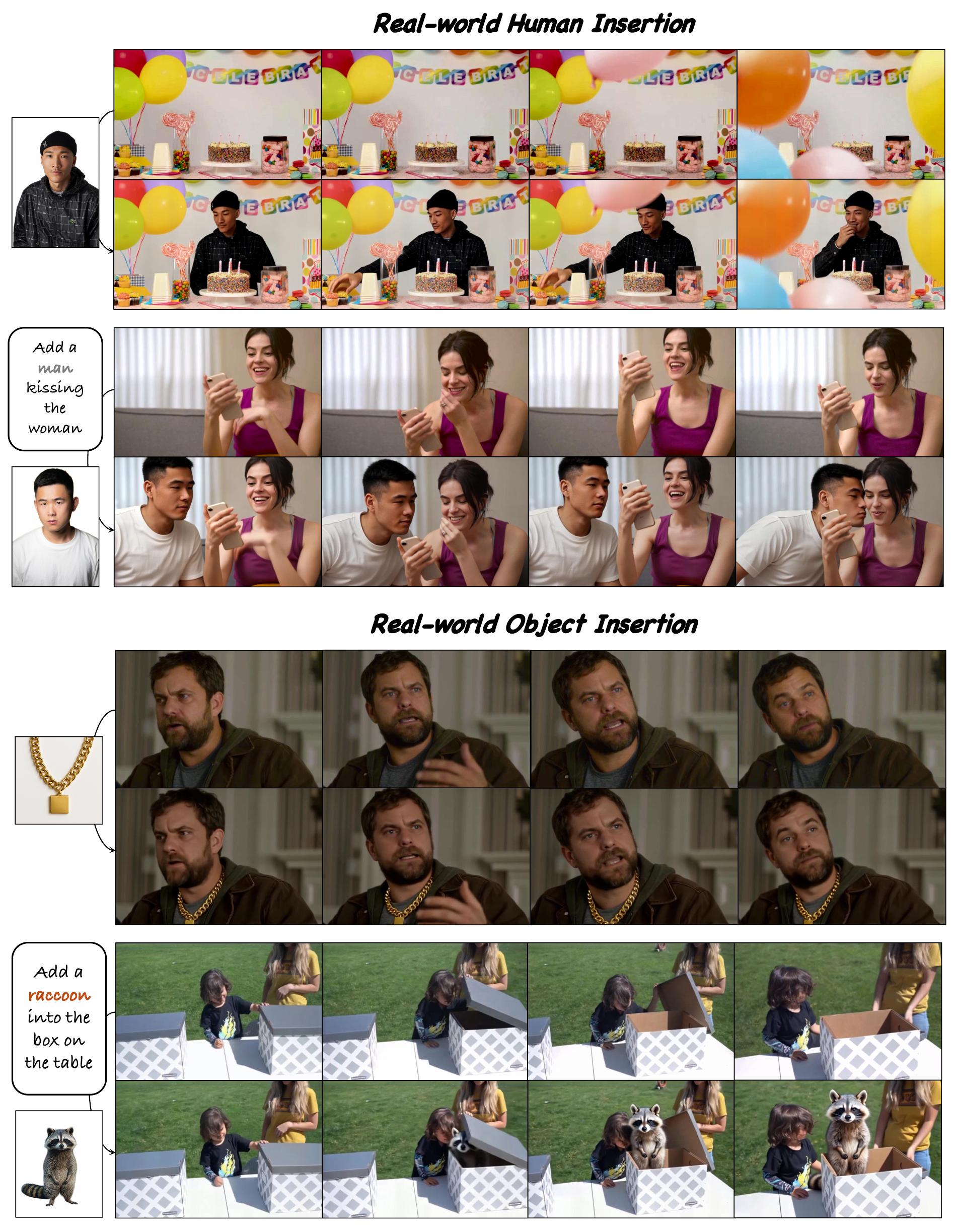}
    \captionof{figure}{Showcase of \textbf{OmniInsert in Real-World scenarios}.}
    \label{fig: teaser1}
\end{figure}

\begin{figure}[H]
    \centering	
    \includegraphics[width=\textwidth]{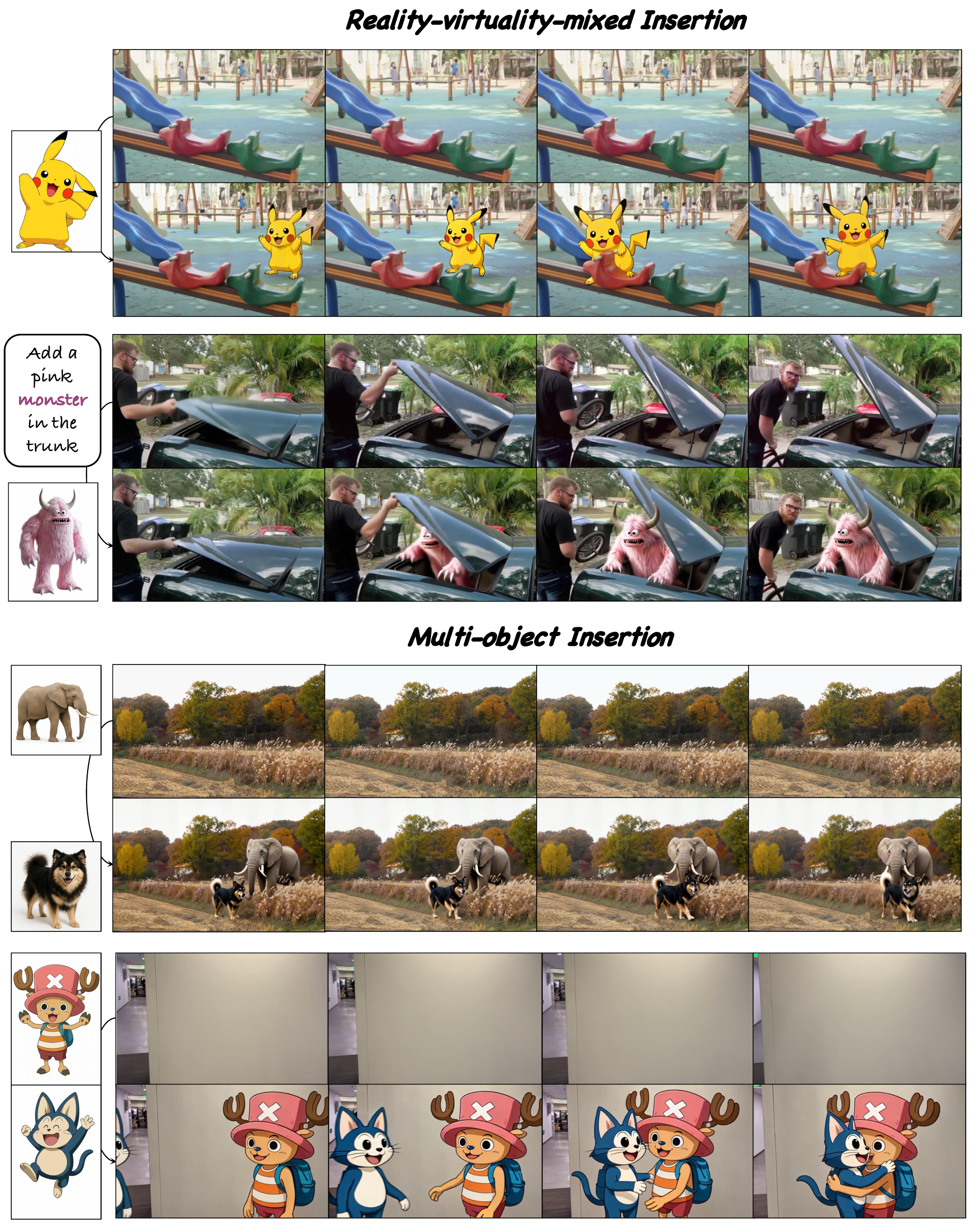}
    \captionof{figure}{Showcase of \textbf{OmniInsert in Reality-virtuality-mixed and Multi-object scenarios}.}
    \label{fig: teaser2}
    \vspace{3mm}
\end{figure}

Prior approaches~\cite{ku2024anyv2v, zhao2023make} employ DDIM inversion for noise initialization to maintain the primary structure of the reference video and inject subject features during the denoising steps. However, these methods introduce an extra inference stage that doubles the computational time and frequently cause discontinuity and unnaturalness in the inserted subject. Recently, VideoAnyDoor~\cite{tu2025videoanydoor} and GetInVideo~\cite{zhuang2025get} propose end-to-end video insertion frameworks to process reference images and condition videos simultaneously. Nevertheless, they rely on complex control signals (e.g., masks or points) to guide insertion position and motion and struggle with subject consistency. Very recently, some unified video editing works~\cite{liang2025omniv2v, ye2025unic} support video insertion based on instruction. However, they still suffer from issues of subject inconsistency and unnatural interactions, limiting practical applicability.

In this work, our focus is on the task of \textit{\textbf{Mask-free Video Insertion (MVI)}}, inserting user-defined characters into a reference video according to the customized prompt. The task poses several challenges, including: 1) Data Scarcity: Lacking paired videos before and after insertion along with corresponding subject references; 2) Subject-scene Equilibrium: Ensuring consistency of inserted subject while maintaining invariance of unedited areas of the reference video; 3) Insertion Harmonization: Achieving plausible positioning and motion of inserted subjects to ensure natural interactions.

To address the Data Scarcity, we propose a new data pipeline \textit{\textbf{InsertPipe}}, producing training data consisting of reference subjects paired with appropriately edited videos and textual prompt. Specifically, we introduce three data curation pipelines: RealCapture Pipe, SynthGen Pipe, and SimInteract Pipe, aiming to enhance data diversity, thereby improving the model’s robustness across complex scenarios. 
The RealCapture Pipe leverages existing real-world videos, constructing paired videos through detection, tracking, and video erasing tools. Unlike prior methods~\cite{huang2025conceptmaster, chen2025multi} that extract key frames from videos as reference subjects, we focus on constructing cross-video subject pairs to prevent copy-paste issue. To ensure comprehensive scene diversity, the SynthGen Pipe utilizes Large Language Model (LLM)~\cite{GPT4} to generate varied prompts. These prompts are combined with techniques like image generation, image editing, video generation, and subject removal to automatically construct large-scale cross-pair datasets. For the scarcity of complex scene interaction data, we design the SimInteract Pipe to generate customized data based on the rendering engine. 

Building upon our data pipeline, we develop \textit{\textbf{OmniInsert}}, a novel unified framework for mask-free video insertion, supporting both single and multiple subject references with customized prompts. To maintain Subject-scene Equilibrium, we introduce three targeted improvements across model architecture, training strategy, and supervision. Firstly, we introduce a simple yet effective Condition-Specific Feature Injection (CFI) mechanism. It distinctly injects conditions from the source video and reference subject while preserving efficiency, requiring only minimal adjustments to the video foundation model. Secondly, we propose a novel Progressive Training (PT) strategy, which enables the model to balance multi-condition injection through multi-stage optimization. Thirdly, we design a Subject-Focused Loss (SL) to aid the model in focusing on capturing the detailed appearance of the subjects. 

To enhance Insertion Harmonization in complex visual scenarios, we propose two complementary techniques. First, Insertive Preference Optimization (IPO) guides the model to learn context-aware insertion strategies using a curated set of paired videos that reflect human preferences across diverse scenes. Second, the Context-Aware Rephraser (CAR) enriches user prompts at inference time by injecting fine-grained scene details (such as object textures, spatial layout, and interaction cues) into the instruction. This allows inserted content to better align with both the semantics and visual structure of the scene.


Through the aforementioned improvements, our approach enables effective insertion of customized subjects into reference videos, as illustrated in Fig.~\ref{fig: teaser1} and Fig.~\ref{fig: teaser2}. To address the lack of a benchmark for MVI, we introduce a comprehensive benchmark, \textit{\textbf{InsertBench}}, which consists of 120 videos paired with meticulously selected subjects (suitable for insertion in each video) and the corresponding prompts. Leveraging this benchmark, we conduct extensive evaluations demonstrating that OmniInsert achieves clear advantages over state-of-the-art commercial solutions, both quantitatively and qualitatively. Overall, our contributions are summarized as follows.

\begin{figure*}[t]
    \centering
    \includegraphics[width=\textwidth]{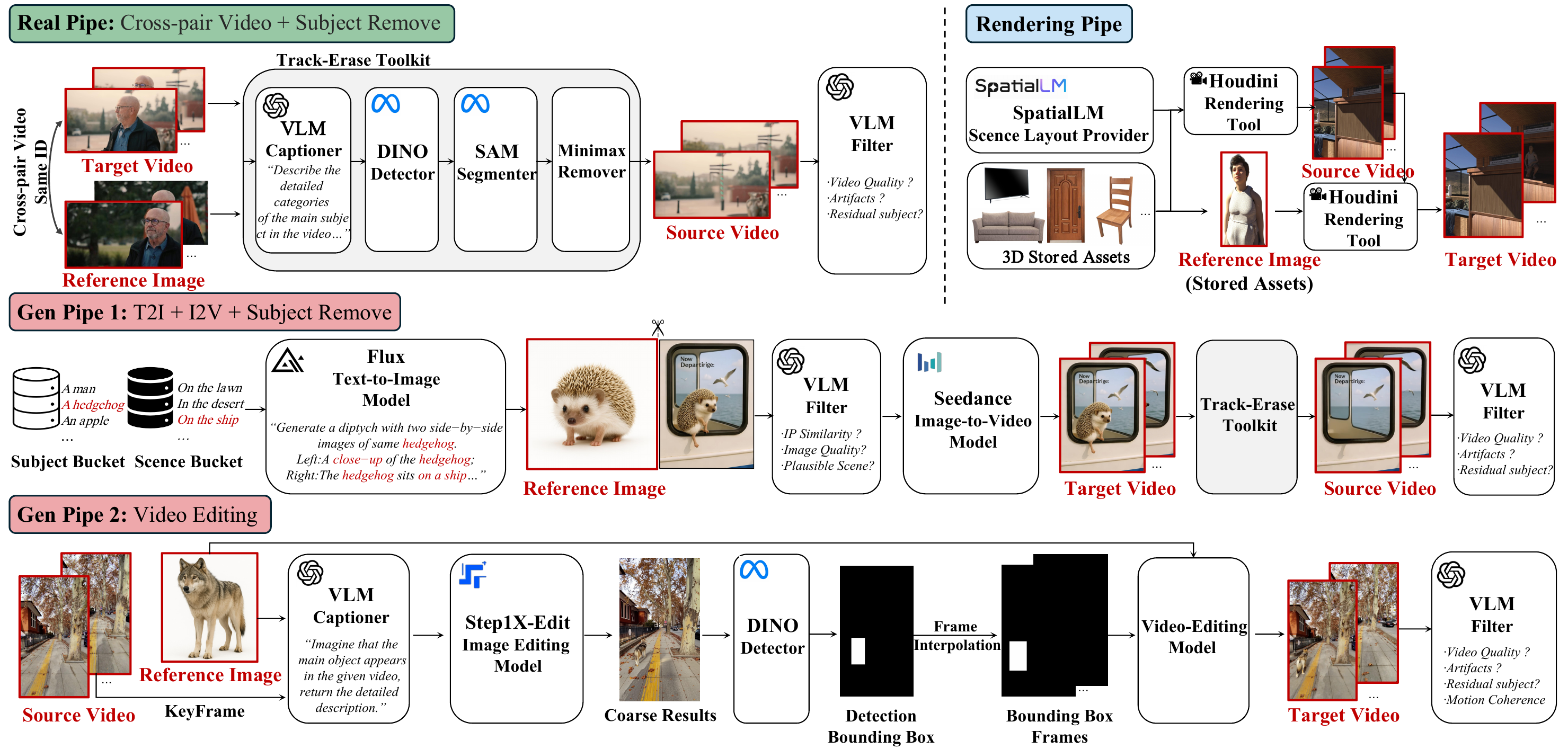}
    \caption{\textbf{Overview of \textit{InsertPipe}.} It consists of three data construction pipelines: Real Pipe, Rendering Pipe, and Gen Pipe.}
    \label{fig: pipes}
\end{figure*}

\textbf{Technology.} 1) We develop \textit{InsertPipe}, a systematic data curation framework featuring multiple data pipelines to automatically generate high-quality and diverse data; 2) We propose \textit{OmniInsert}, a unified mask-free architecture capable of seamlessly inserting both single and multiple reference subjects into videos; 3) We introduce \textit{InsertBench}, a comprehensive benchmark tailored for MVI task.

\textbf{Significance.} 1) \textit{OmniInsert} demonstrates superior generation quality, bridging the gap between academic research and commercial-grade applications; 2) We present a comprehensive study of the MVI task—including data, model, and benchmark—will be publicly released to support future research and development.

%% file: sec/3_related.tex
\section{Related Work}
\noindent\textbf{Video Foundation Model.}
The development of diffusion models~\cite{ho2020denoising} has significantly advanced video foundation model research. Early models~\cite{blattmann2023stable, guo2023animatediff} extended pre-trained Text-to-Image (T2I) models for continuous video generation by adding temporal modules. VDM~\cite{ho2022video} extends 2D U-Net to 3D, while AnimateDiff~\cite{guo2023animatediff} integrates 1D temporal attention into 2D spatial blocks for efficiency. These strategies allow them to leverage powerful image priors, but their capabilities can be constrained by the original image-centric architecture. Recently, the emergence of Diffusion Transformer(DiT)~\cite{peebles2023scalable}-based methods for video generation has exhibited superior performance in quality and consistency. These methods treat video as a sequence of spatiotemporal patches, processing them in a unified manner with a Transformer. These powerful scaling transformers~\cite{liu2024sora, wan2025wan, gao2025seedance, kong2024hunyuanvideo, yang2024cogvideox, ma2025step} enable generating longer and higher-quality videos, paving the way for various downstream video generation tasks.

\noindent\textbf{Video Insertion.}
Video Insertion (VI) aims to insert a user-provided reference subject into source video while maintaining the consistency of unedited regions. Prior approaches~\cite{ku2024anyv2v, zhao2023make} employ DDIM inversion to initialize the generation noise based on the reference video and inject subject features during the denoising steps. Due to they inherently require an additional stage for inversion, computational overhead and inference costs increase significantly. VideoAnyDoor~\cite{tu2025videoanydoor} proposes an end-to-end framework that precisely modifies both content and motion according to user-specified mask and point conditions. GetInVideo~\cite{zhuang2025get} employs a 3D full-attention diffusion transformer architecture to jointly process reference images, condition videos, and masks, maintaining temporal coherence. However, these methods rely on complex control signals but struggle with subject consistency. Recently, some unified video editing works~\cite{liang2025omniv2v, ye2025unic} support mask-free video insertion but still encounter subject infidelity and incoherent interactions in complex scenarios. Consider commercial software~\cite{Pika, Keling} currently maintains state-of-the-art in MVI capabilities, it's critical to bridge the gap between academic research and commercialization.

%% file: sec/4_methods.tex
\section{Preliminary}
The Diffusion Transformer (DiT)~\cite{peebles2023scalable} model employs a transformer as the denoising network to refine diffusion latent. Our method inherits the video diffusion transformers trained using Flow Matching~\cite{lipman2022flow}, which conducts the forward process by linearly interpolating between noise and data in a straight line. At the time step $t$, latent $\mathbf{z}_t$ is defined as: $\mathbf{z}_t = (1-t)\mathbf{z}_0 + t\epsilon$, where $\mathbf{z}_0$ is the clean video, and $\epsilon \sim \mathcal{N}(0, 1)$ is the Gaussian noise. The model is trained to directly regress the target velocity:
\begin{equation}
    \mathcal{L}_{\text{FM}} = \mathbb{E}[\|(\mathbf{z}_0 - \epsilon) - \mathbf{\mathbf{v}}_\theta(\mathbf{z}_t, t, y)\|^2],
    \label{eq:placeholder}
\end{equation}
where $\mathbf{\mathbf{v}}_\theta$ refers to the diffusion model output and $y$ denotes condition. Our diffusion model is composed of stacked DiT blocks, which perform 2D self-attention for spatial modeling and 3D self-attention for spatio-temporal fusion.

\section{Methodology}
Given reference subjects and a source video, we aim to generate a new composited video showcasing a natural interaction between the inserted contents and the original scenes. We first introduce a new data pipeline \textit{InsertPipe} to construct diverse paired data (Sec.~\ref{subsec: InsertPipe}). Building upon our data pipeline, we develop a novel framework \textit{OmniInsert} employing a core Condition-Specific Feature Injection (CFI) module for mask-free video insertion (Sec.~\ref{subsec: Framework}). To further enhance insertion stability and quality, we design a Progressive Training (PT) strategy incorporating the Subject-Focused Loss (SL) and Insertive Preference Optimization (IPO) during training (Sec.~\ref{subsec: Training}). Additionally, we propose a Context-Aware Rephraser (CAR) module during inference to strengthen insertion harmonization (Sec.~\ref{subsec: Inference}). 

\subsection{InsertPipe} \label{subsec: InsertPipe}
To address the challenge of data scarcity, we introduce a comprehensive data pipeline \textit{InsertPipe} to produce diverse data for the MVI task, encompassing RealCapture Pipe, SynthGen Pipe, and SimInteract Pipe, as shown in Fig.~\ref{fig: pipes}. The resulting training samples are formatted as \{\textit{prompt}, \textit{reference images}, \textit{source video}, \textit{target video}\}.

\noindent\textbf{RealCapture Pipe.}
Using long videos from~\cite{wang2025koala} and proprietary data, we segment videos into single-scene clips as target videos with AutoShot/PySceneDetect. The Vision-Language Model (VLM)~\cite{GPT4} then captions these clips, detailing subject appearance, scenes, and interactions. These captions are processed by LLM~\cite{GPT4} to extract subject categories/appearances for detection/tracking prompts~\cite{ravi2024sam2}, generating mask sequences. Furthermore, we apply video erasing techniques~\cite{zi2025minimax} to remove target subjects to create source videos, with a VLM-based filtering mechanism that avoids visual artifacts of inpainting. For subject conditions, we construct cross-video pairs (avoiding same-video keyframes~\cite{chen2025multi,huang2025conceptmaster}) to prevent copy-paste issues. Subjects are matched across CLIP~\cite{shao20221st} and facial embeddings~\cite{deng2019arcface}, discarding pairs with extreme similarity (copy-paste) or dissimilarity (inconsistency).

\noindent\textbf{SynthGen Pipe.}
Considering the limited scene diversity, we introduce a generation-based data pipeline comprising two complementary sub-flows. 1) We first meticulously construct subject and scene buckets with 300 categories and 1000 settings, respectively. Then, diverse subject-scene pairs generated by LLM are combined with text templates to produce cross-pair references via T2I models~\cite{flux2024, labs2025flux1kontextflowmatching}. 
To ensure consistency, VLM is further employed to score the appearance consistency and detail preservation within each pair. 
Subsequently, we synthesize target videos depicting natural interactions using the Image-to-Video (I2V) foundation models~\cite{gao2025seedance}, while source videos are derived using the tracking-erase toolkit with the VLM filtering strategy. 2) Given videos depict empty scenes, we employ VLM to determine suitable inserted subjects and generate corresponding prompts. Next, we sample key frames and apply instruction-based image editing methods~\cite{liu2025step1x} to get target coarse positions. Leveraging temporal interpolation and video inpainting to synthesize the target videos, with VLM filtering ensuring temporal coherence and subject consistency.

\begin{figure*}[t]
    \begin{center}
    \includegraphics[width=0.8\textwidth]{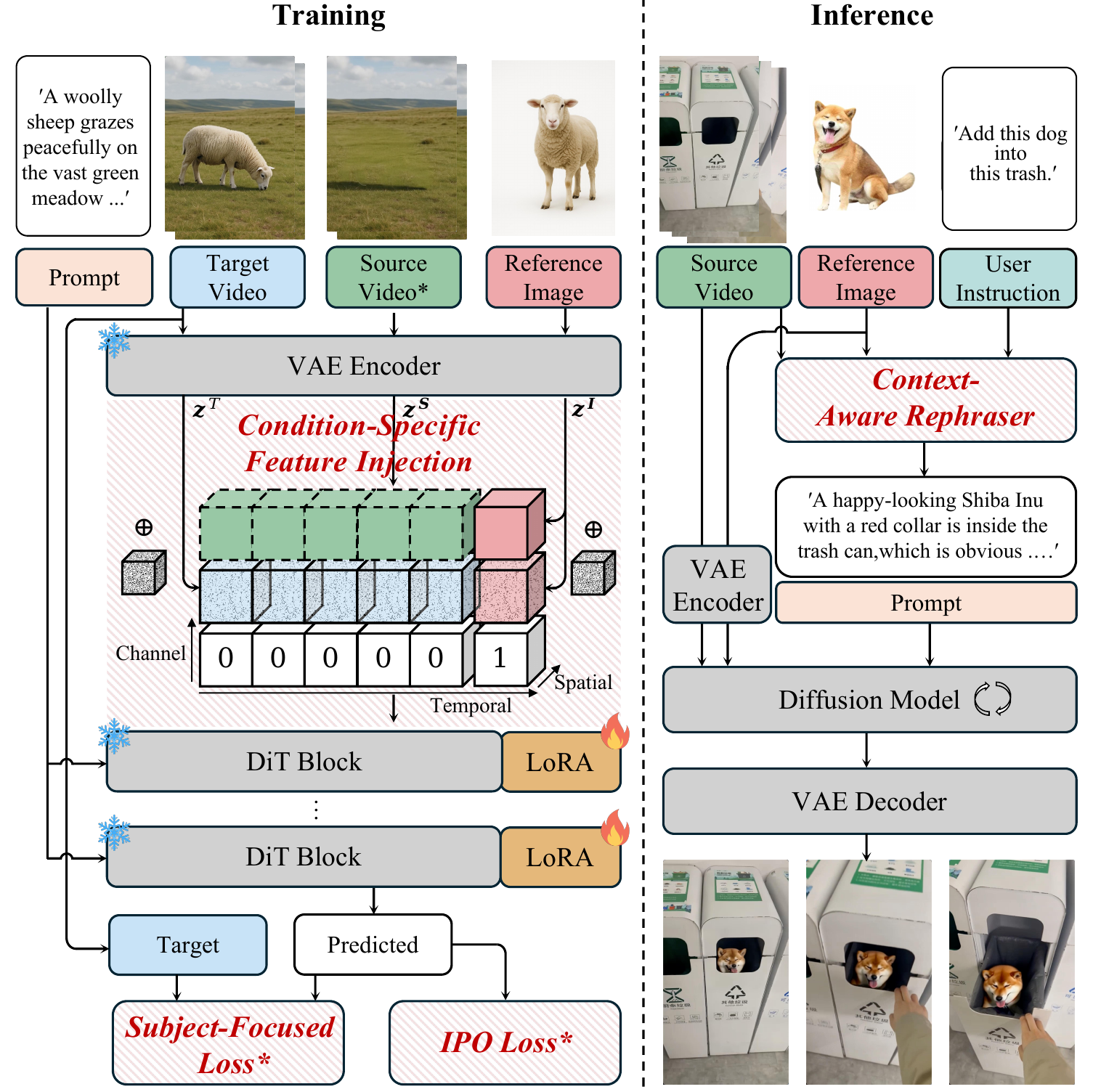}
    \end{center}
    \caption{\textbf{Overview of \textit{OmniInsert}.} 
    Note that the parts marked with * are only enabled in specific phases.}
    \label{fig: framework}
\end{figure*}

\noindent\textbf{SimInteract Pipe.}
Despite enhanced diversity, complex subject-scene interactions (e.g., a man waving behind slowly opening doors) remain scarce. To address this, we develop a rendering pipeline based on~\cite{Houdini}. Specifically, we construct extensive asset libraries for subjects and scenes in the rendering environment. SpatialLM~\cite{mao2025spatiallm} then generates 350 layout priors to drive stochastic asset placement. Leveraging rigged assets with predefined motion bindings, we synthesize interactions by retrieving motion library animations through predefined prompts. Finally, we can leverage strategically positioned cameras to render target videos while paired source videos are derived via subject removal under identical cameras.

\subsection{OmniInsert Framework} \label{subsec: Framework}
Mask-free Video insertion is a challenging task that requires accurate preservation of both subject identity and background consistency. A straightforward solution is to inject VAE features of the references along the temporal dimension or to concatenate reference visual tokens after patchification. However, such approaches impose high computational overhead, and fail to account for the distinct alignment requirements of different conditions: reference videos demand frame-wise alignment with latent noise, while subject references require full temporal feature interaction. 

To address these challenges, we propose \textit{OmniInsert}, a novel end-to-end framework for mask-free video insertion using single or multiple subject references guided by text prompts, as shown in Fig.~\ref{fig: framework}. At its core, we introduce a simple yet effective Condition-Specific Feature Injection (CFI) mechanism that injects both video and subject features in a unified yet efficient manner, tailored to their respective temporal alignment needs.


\noindent\textbf{Condition-Specific Feature Injection.}
CFI adopts targeted designs in both the injection mechanism and channel-level flags to accommodate the distinct characteristics of video and object conditions.
Specifically, for the video condition, source videos are concatenated with latent noise along the channel dimension, accompanied by dedicated flags, formally expressed as:
\begin{equation}
    \mathbf{z}^{\text{Vid}}_{t} = \text{Concat}([\mathbf{z}^{T}_t, \mathbf{z}^S, \mathbf{f}^S], \text{dim}=1), 
\end{equation}
where noisy target video latent $\mathbf{z}^{T}_{t} = (1-t)\mathbf{z}^{T} + t\epsilon$, $\mathbf{z}^{T} \in \mathbb{R}^{f \times c \times h \times w}$ denotes the clean target video latent. 
$\mathbf{z}^S \in\mathbb{R}^{f\times c\times h\times w}$ represents reference source video latent, $\mathbf{f}^{S} \in \mathbb{R}^{f \times 1 \times h \times w}$ is an all-zero vector. 
The choice of channel-wise concatenation aligns well with the background condition, facilitating effective spatial alignment.

In contrast, for the subject condition, latents are concatenated along the temporal dimension to better capture dynamic changes across frames. As shown in Fig.~\ref{fig: framework}, the subject latent is constructed as:
\begin{equation}
\mathbf{z}^{\text{Sub}}_{t} = \text{Concat}([\mathbf{z}^{I}_{t}, \mathbf{z}^I, \mathbf{f}^I], \text{dim}=1),
\end{equation}
where $\mathbf{z}^{I}, \mathbf{z}^{I}_{t} \in \mathbb{R}^{n \times c \times h \times w}$ denote the clean and noisy subject latents, with $\mathbf{z}^{I}_{t} = (1 - t)\mathbf{z}^{I} + t\epsilon$, and $n$ denotes the number of inserted subjects.
$\mathbf{f}^I \in \mathbb{R}^{1 \times 1 \times h \times w}$ is an all-one flag map.
The overall diffusion input is finally constructed by concatenating the video and subject condition latents along the frame dimension: $\mathbf{z}_t = \text{Concat}([\mathbf{z}^{\text{Vid}}_t, \mathbf{z}^{\text{Sub}}_t], \text{dim} = 0)$. This temporal concatenation effectively captures subject motion and continuity.

Thanks to our differentiated condition injection strategies, CFI effectively achieves seamless subject insertion and background preservation while maintaining network efficiency. Furthermore, we integrate the LoRA mechanism into the DiT blocks to avoid expensive full-parameter updates, thereby preserving the model’s original text-alignment capabilities and visual fidelity.

\subsection{OmniInsert Training Pipeline} \label{subsec: Training}
\noindent\textbf{Progressive Training.} 
A significant challenge in training video insertion models lies in the inherent imbalance of learning difficulty: preserving the source video is mainly a copy-paste task, which is much simpler than the complex temporal and spatial modeling involved in dynamic subject insertion. 
Consequently, naive single-stage training tends to bias the model toward the simpler task of background preservation, often resulting in failed object insertion, as illustrated in Fig~\ref{fig: ablation}. 
To mitigate these issues, we introduce a four-stage progressive training strategy: 

Phase 1. We begin by training the model to inject reference subjects based on subject features and text prompts, while discarding the source video condition. This isolates subject modeling and helps the model develop strong subject representation and motion generation capabilities.

Phase 2. Based on the phase 1 model, we introduce the source video and conduct pretraining on the complete MVI task across the full dataset. This enables the model to acquire an initial capacity for aligning subjects with video backgrounds. However, models at this stage often suffer from identity inconsistency and show limited performance in complex scenes.

Phase 3. To address the above limitations, we conduct a refinement stage using a curated dataset composed of high-fidelity portraits and synthetic renderings. A short period of fine-tuning on this data significantly enhances identity preservation and improves the model’s robustness in visually complex environments.

Phase 4. To further reduce physical implausibilities (e.g., unnatural poses, model penetration) and visual artifacts, we develop Insertion Preference Optimization (IPO), a fine-tuning stage inspired by~\cite{rafailov2023direct}. 
By leveraging a small number of human-annotated preference pairs, IPO guides the model toward more visually and physically plausible insertions, yielding notable improvements in generation quality.



Through progressive optimization, the model achieves balanced feature injection from subjects and source video, enhancing insertion stability and fidelity. 
Notably, the success of this progressive strategy also critically depends on carefully designed losses, which are described below.

\noindent\textbf{Subject-Focused Loss.} 
In Phases 1–3, the flow matching loss ($\mathcal{L}_{\text{FM}}$) serves as the primary training objective. However, since it treats all regions of the synthesized video equally, it often fails to maintain consistency for subjects that occupy relatively small spatial areas.
To address this issue, we propose the Subject-Focused Loss (SL), a modification of $\mathcal{L}_{\text{FM}}$ that directs the model’s attention more strongly toward the subject regions. Formally, the loss is defined as:
\begin{equation}
    \mathcal{L}_{\text{SL}} = \mathbb{E}[\| \mathcal{M} \cdot [(\mathbf{z}_0 - \epsilon) - \mathbf{\mathcal{V}}_\theta(\mathbf{z}_t, t, y)]\|^2],
\end{equation}
where $\mathcal{M}$ denotes spatially downsampled masks derived from tracking. Thus, the overall loss $\mathcal{L}$ for Phase 1-3 is:
\begin{equation}
    \mathcal{L} = \lambda_1 \mathcal{L}_{\text{FM}} + \lambda_2 \mathcal{L}_{\text{SL}}.
\end{equation}

\input{tables/comparision}
\input{tables/user_study}

\noindent\textbf{IPO Loss.}
In Phase 4, we propose a preference optimization mechanism to improve visual stability and physical plausibility. Specifically, the well-trained phase-3 model serves as the reference model $\pi_{\text{ref}}$, while an additional LoRA module is initialized as the trainable policy $\pi_\theta$. For preference data construction, we curate paired samples (preferred $y_w$ and dispreferred $y_l$) through human selection of model inference outputs. Similar to~\cite{xiao2024cal}, the loss is:
\begin{equation}
    \mathcal{L}_{\text{IPO}} = \mathcal{L}_{\text{DPO}} + \lambda \cdot \mathbb{E} \left[ \left( -\log \pi_\theta(y_l|x) - \gamma \right)^2 \right], 
    \label{eq:ipo_loss}
\end{equation}
\begin{small}
\begin{equation}
    \mathcal{L}_{\text{DPO}} = -\mathbb{E}\left[ \log \sigma \left( \beta \log \frac{\pi_\theta(y_w|x)}{\pi_{\text{ref}}(y_w|x)} - \beta \log \frac{\pi_\theta(y_l|x)}{\pi_{\text{ref}}(y_l|x)} \right) \right], 
    \label{eq:dpo_loss}
\end{equation}
\end{small}
where $x$ denotes the model input for brevity and $\sigma$ represents the standard Sigmoid function. Notably, we find that a limited dataset of only 500 preference pairs suffices to achieve substantial performance gains at this stage. Note that the IPO loss is exclusively applied in phase 4 as the sole objective function.

\subsection{OmniInsert Inference Pipeline} \label{subsec: Inference}
Classifier-free guidance~\cite{ho2022classifier} balances sample quality and diversity in diffusion models through joint conditional and unconditional training. During inference, we design a joint classifier-free guidance to balance multiple conditions:
\begin{small} 
\begin{equation}
    \begin{aligned}
       \hat{\mathbf{v}}_{\theta}(\mathbf{z}_t, \mathbf{c}_p, \mathbf{c}_i, \mathbf{c}_v) &= \mathbf{v}_{\theta}(\mathbf{z}_t, \emptyset, \emptyset, \emptyset) \\
       &+ S_1 \cdot \left( \mathbf{v}_{\theta}(\mathbf{z}_t, \mathbf{c}_p, \emptyset, \emptyset) - \mathbf{v}_{\theta}(\mathbf{z}_t, \emptyset, \emptyset, \emptyset) \right) \\
       &+ S_2 \cdot \left( \mathbf{v}_{\theta}(\mathbf{z}_t, \mathbf{c}_p, \mathbf{c}_i, \emptyset) - \mathbf{v}_{\theta}(\mathbf{z}_t, \mathbf{c}_p, \emptyset, \emptyset) \right) \\
       &+ S_3 \cdot \left( \mathbf{v}_{\theta}(\mathbf{z}_t, \mathbf{c}_p, \mathbf{c}_i, \mathbf{c}_v) - \mathbf{v}_{\theta}(\mathbf{z}_t, \mathbf{c}_p, \mathbf{c}_i, \emptyset) \right),
    \end{aligned}
\end{equation}
\end{small}
where $\mathbf{c}_p$, $\mathbf{c}_i$, $\mathbf{c}_v$ represent prompt condition, reference subjects, and source video. $S_1$, $S_2$ and $S_3$ are guidance scales.

\noindent\textbf{Context-Aware Rephraser.} 
To enhance the coherence of subject insertion in complex visual scenes, we introduce a Context-Aware Rephraser (CAR) module at inference time, as shown in Fig~\ref{fig: framework}. CAR leverages the VLM to generate detailed, context-aware prompts that guide the model toward more seamless and plausible integration of the inserted subject into the source scene. Specifically, CAR first prompts the VLM to produce fine-grained descriptions of both the source environment and the reference subject. It then provides the VLM with creative constraints—such as preserving the original video setting, encouraging imaginative yet coherent insertion of the subject, and maintaining spatial consistency in terms of size and position, etc. As a result, CAR produces prompts that more accurately reflect the user's creative intent, particularly in terms of visual effects (VFX), ultimately improving the realism and harmony of insertions achieved by \textit{OmniInsert}.

%% file: tables/comparision.tex
\setlength{\tabcolsep}{0.55mm}{
\begin{table*}[t]
\small
\centering
\renewcommand{\arraystretch}{1.0}
\resizebox{\textwidth}{!}{
\begin{tabular}{l|ccc|c|cccc} 
\Xhline{1pt}
\multirow{2}{*}{Method} & \multicolumn{3}{c|}{\textbf{Subject Consistency}} & \textbf{Text-Video Alignment} & \multicolumn{4}{c}{\textbf{Video Quality}}\\
 \cline{2-9}
 & {CLIP-I$^*$ $\uparrow$} & {DINO-I$^*$} $\uparrow$ & FaceSim $\uparrow$ & ViCLIP-T $\uparrow$ & Dynamic $\uparrow$ & Image-Quality $\uparrow$ & Aesthetics $\uparrow$  & Consistency $\uparrow$\\
\midrule
Pika-Pro~\cite{Pika}   & \underline{0.682} & \underline{0.543} & 0.422 & \underline{24.721} & \textbf{0.873} & 0.655 & \underline{0.524} & 0.910 \\
Kling~\cite{Keling}   & 0.664 & 0.513 & \textbf{0.539} & 23.091 & 0.823 & \underline{0.667} & 0.521 & \underline{0.921}  \\
\rowcolor{myblue} $\textbf{Ours}$ & \textbf{0.745} & \textbf{0.639} & \underline{0.488} & \textbf{25.945} & \underline{0.825} & \textbf{0.704} & \textbf{0.556} & \textbf{0.930} \\
\Xhline{1pt}
\end{tabular}
}
\vspace{-1mm}
\caption{ \textbf{Quantitative comparisons} with baseline methods, and $^*$ indicates the consistency of the segmented subject area.
}
\vspace{-2mm}
\label{tab: comp_single}
\end{table*}
}

%% file: tables/user_study.tex
\setlength{\tabcolsep}{0.3mm}{
\begin{table}[t]
\small
\centering
\renewcommand{\arraystretch}{0.85}
\resizebox{0.65\textwidth}{!}{
\begin{tabular}{l|cccc}
\Xhline{1pt}
        \small Method & \begin{tabular}[c]{@{}c@{}} \scriptsize Subject \vspace{-2pt} \\ \scriptsize Consistency $\uparrow$ \end{tabular} & \begin{tabular}[c]{@{}c@{}} \scriptsize Align with \vspace{-2pt} \\ \scriptsize Prompt $\uparrow$ \end{tabular} & \begin{tabular}[c]{@{}c@{}} \scriptsize Insertion \vspace{-2pt} \\ \scriptsize Rationality $\uparrow$ \end{tabular} & \begin{tabular}[c]{@{}c@{}} \scriptsize Comprehensive \vspace{-2pt} \\ \scriptsize Evaluation $\uparrow$ \end{tabular} \\ 
\midrule
Pika-Pro~\cite{Pika} & 5.08\% & 9.00\% & 10.67\% & 8.58\% \\
Kling~\cite{Keling} & 29.42\% & 22.67\% & 24.92\% & 23.08\% \\
\rowcolor{myblue} \textbf{Ours} & \textbf{65.50\%} & \textbf{68.33\%} & \textbf{64.41\%} & \textbf{68.34\%} \\ 
\Xhline{1pt}
\end{tabular}
}
\caption{\small \textbf{User study} with baseline methods.}
\label{tab: user_study}
\end{table}
}

%% file: sec/5_exps.tex
\section{Experiments}
\begin{figure*}[!ht]
    \centering
    \includegraphics[width=0.75\textwidth]{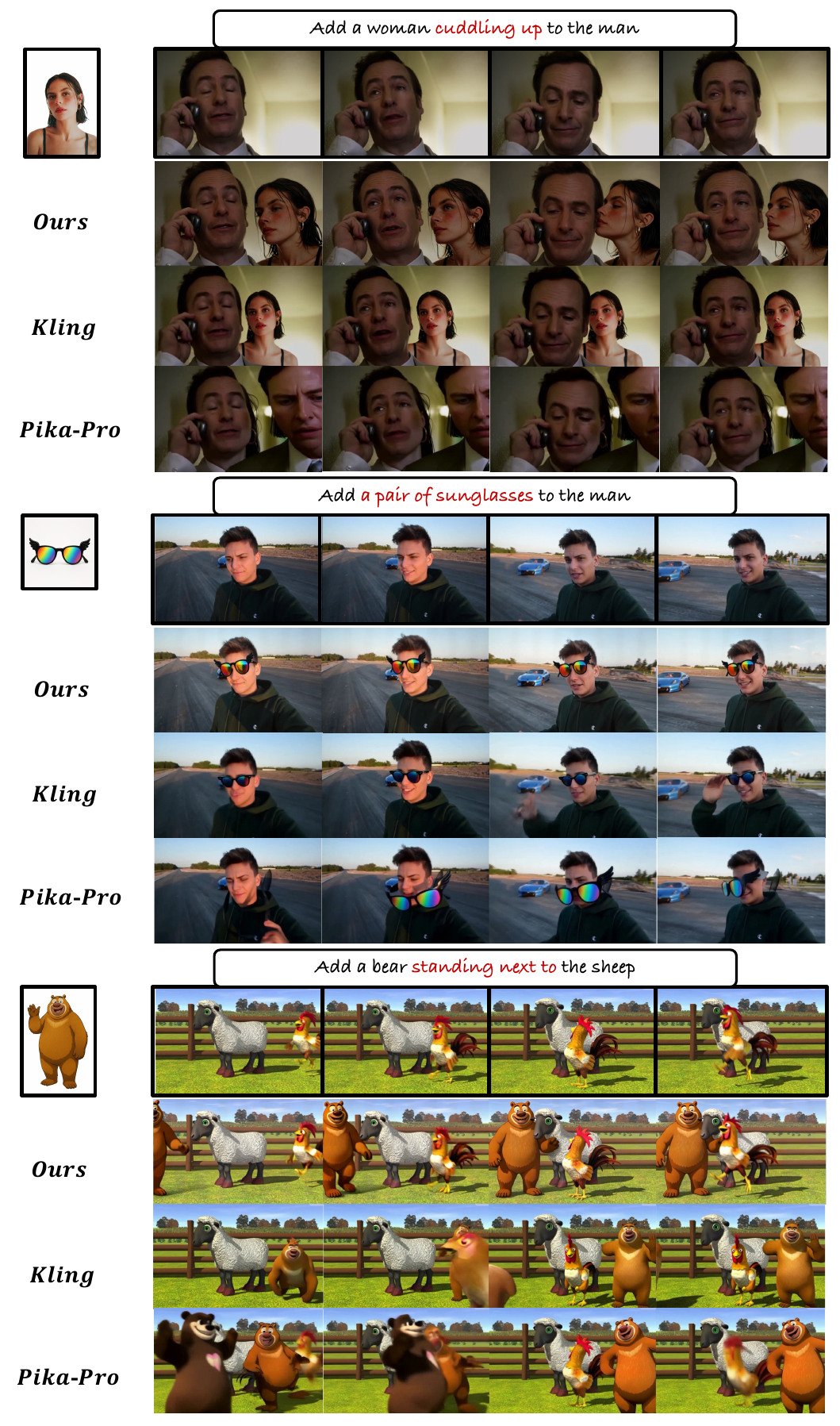}
    \caption{\textbf{Qualitative comparisons I} with state-of-the-art methods. Please zoom in for more details.}
    \label{fig: comparison1}
\end{figure*}
\begin{figure*}[!ht]
    \centering
    \includegraphics[width=0.75\textwidth]{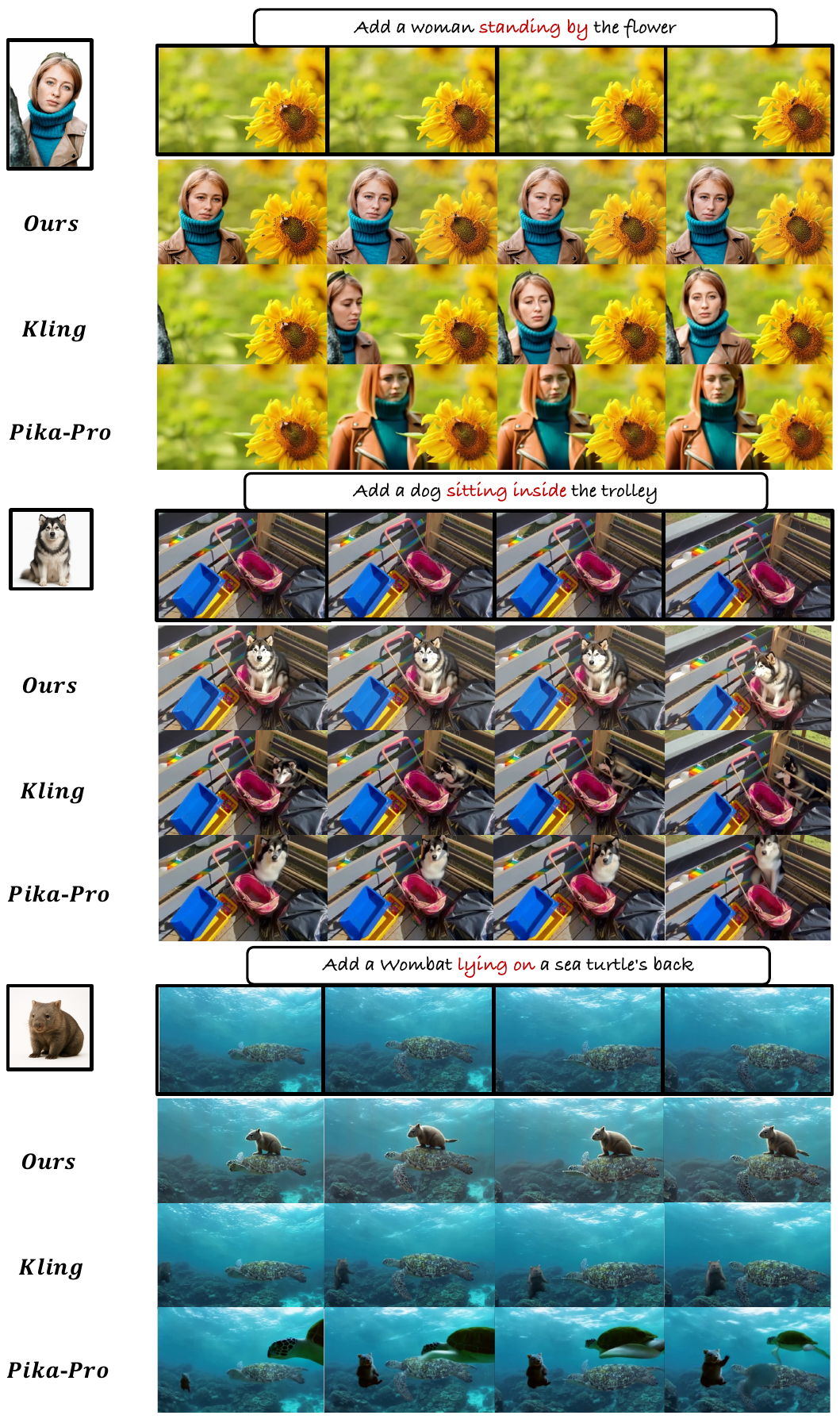}
    \caption{\textbf{Qualitative comparisons II} with state-of-the-art methods. Please zoom in for more details.}
    \label{fig: comparison2}
\end{figure*}
\subsection{Setup}
\noindent\textbf{InsertBench.}
We present \textit{InsertBench}, a comprehensive benchmark for evaluating MVI capabilities. The benchmark comprises 120 videos spanning various environments, including indoor, natural landscapes, wearable scenarios, and even animated scenes. Each video is approximately 5 seconds long with 121 frames, paired with meticulously selected subjects to ensure visually compatible insertion. Such a diverse benchmark would be beneficial for the community.

\noindent\textbf{Implementation Details.} The training protocol comprises four phases: 1) 70k iterations for subject-to-video task in phase 1; 2) 30k iterations for MVI task pretraining in phase 2; 3) 10k iterations for model refinement in phase 3; 4) 8k iterations for preference optimization in phase 4. The total computational resources consumed approximately 7000 GPU-hours on A100. During inference, we use the Euler sampler with 50 steps and set $S_1, S_2, S_3$ to 7.5, 3.5 and 1.5, respectively. For fair comparison, all experiments are conducted at the resolution of 480p. Please refer to the supplementary materials for more details.

\noindent\textbf{Baselines.}
For the MVI task, the existing available state-of-the-art methods are closed-source commercial software. Therefore, we evaluate and compare the latest capabilities of Pika-Pro~\cite{Pika} and Kling~\cite{Keling}.

\noindent\textbf{Evaluation Metrics.}
We assess three dimensions: CLIP-I, DINO-I and FaceSim~\cite{deng2019arcface} for Subject Consistency; ViCLIP-T~\cite{wang2022internvideo} for Text-Video Alignment; Dynamic Quality, Image-Quality, Aesthetics and Consistency~\cite{huang2024vbench++} for Video Quality.

\subsection{Qualitative Analysis}
Fig.~\ref{fig: comparison1} and Fig.~\ref{fig: comparison2} present visual comparisons between our method and baseline approaches. \textit{OmniInsert} demonstrates superior subject/background consistency, whereas baseline methods struggle to balance multi-source feature injection. Specifically, Pika exhibits significant deficiencies in subject fidelity and insertion plausibility, particularly in wearable and human-centric scenarios. Kling suffers from obvious scene inconsistency while encountering copy-paste issues in portrait scenarios. 
Compared with baselines, our method ensures the natural integration of inserted subjects within source scenes. Since the compared methods lack multi-reference support, additional results of multiple subjects are provided in the supplementary materials.

\subsection{Quantitative Comparisons}
\noindent\textbf{Metric Evaluation.} 
For subject consistency, we uniformly sample 10 frames per video and compute their similarity to reference subjects. For prompt following, we measure text-video cosine similarity. Tab.~\ref{tab: comp_single} shows \textit{OmniInsert} leads in metrics of CLIP-I$^*$, DINO-I$^*$ and ViCLIP-T. Although Kling achieves higher face similarity in portrait scenarios, they encounter copy-paste issues (1st line in Fig.~\ref{fig: comparison1} and Fig.~\ref{fig: comparison2}). Moreover, our method shows superiority in video quality. 

\noindent\textbf{User Study.} 
We conduct a user study to evaluate the generation quality of our model. 
We assigned 40 test samples and invited 30 volunteers to select the \textbf{most} preferred result under four criteria: subject consistency, text alignment, insertion rationality and comprehensive evaluation. Tab.~\ref{tab: user_study} shows that \textit{OmniInsert} achieves significant advantages.

\subsection{Ablation Studies}
To demonstrate the effectiveness of our proposed method, we conducted the following ablation studies:
a). Directly training the MVI task on the full dataset without the Progressive Training (PT) method of phases 1-3 (denoted as w/o PT).
b). Not using the Subject-Focused Loss (SL) during training (denoted as w/o SL).
c). Simply using the user instruction without Context-Aware Rephraser (CAR) during inference (denoted as w/o CAR).
d). Not using the Insertion Preference Optimization (IPO) phase during training (denoted as w/o IPO).

\begin{figure}[!ht]
    \begin{center}
    \includegraphics[width=\textwidth]{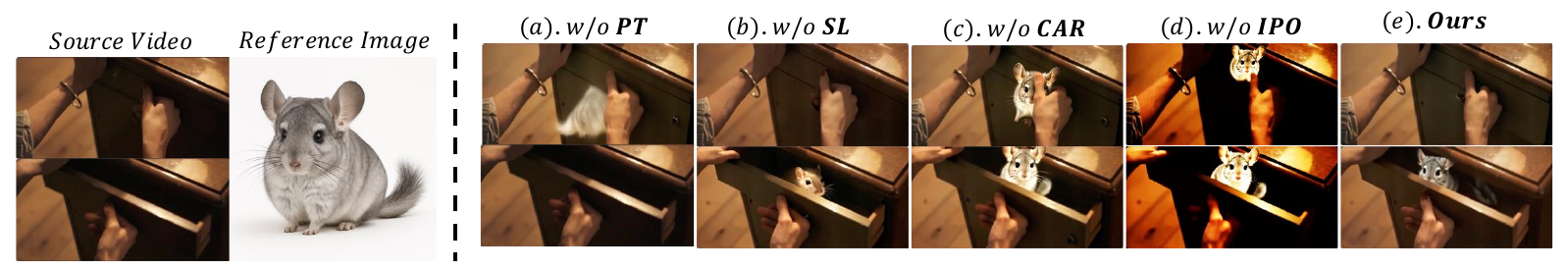}
    \end{center}
    \caption{\textbf{Ablation results} of our method.}
    \label{fig: ablation}
\end{figure}

As shown in Fig.~\ref{fig: ablation} and Table.~\ref{tab: abl}, we demonstrate the effectiveness of our proposed modules:
\textbf{a). PT} substantially facilitates subject-scene equilibrium through multi-stage optimization, enhancing insertion stability; 
\textbf{b). SL} guides the model to focus on more critical yet harder-to-fit subject regions, improving subject preservation; 
\textbf{c). CAR} generates rich descriptions of subject-scene interactions, facilitating more natural compositions; 
\textbf{d). IPO} effectively enhances insertion rationality while mitigating visual artifacts.

\input{tables/ablation}


%% file: tables/ablation.tex

\setlength{\tabcolsep}{0.3mm}{
\begin{table}[ht]
\small
\centering
\renewcommand{\arraystretch}{0.8}
\resizebox{0.55\textwidth}{!}{
\begin{tabular}{l|cccc}
\Xhline{1pt}
\small Method & \footnotesize CLIP-I$^*$ $\uparrow$ & \footnotesize DINO-I$^*$ $\uparrow$ & \footnotesize ViCLIP-T $\uparrow$ & \footnotesize Aesthetic $\uparrow$ \\
\midrule
a) w/o PT & 0.642 & 0.533 & 22.883 & 0.520 \\
b) w/o SL & 0.657 & 0.587 & 25.908 & 0.542 \\
c) w/o CAR & 0.689 & 0.609 & 23.052 & 0.532 \\
d) w/o IPO & 0.732 & 0.625 & 24.537 & 0.507 \\
\textbf{e) Ours} & \textbf{0.745} & \textbf{0.639} & \textbf{25.945} & \textbf{0.556} \\
\Xhline{1pt}
\end{tabular}
}
\caption{\small \textbf{Ablation study} of our method.}
\label{tab: abl}
\end{table}
}

%% file: sec/6_conclusion.tex
\section{Conclusion} 
This paper presents a comprehensive study of the task, i.e., \textit{\textbf{Mask-free Video Insertion}}. To address the data scarcity, we introduce a new data pipe \textit{\textbf{InsertPipe}} to produce diverse paired data. Furthermore, we develop a novel unified framework \textit{\textbf{OmniInsert}}, leveraging a Condition-Specific Feature Injection mechanism and incorporating a Progressive Training strategy with Subject-Focused Loss to achieve subject-scene equilibrium. Additionally, we propose an Insertive Preference Optimization methodology and a Context-Aware Rephraser module to enhance insertion harmonization. Finally, to address the lack of a benchmark, we build \textit{\textbf{InsertBench}} and experiments show that \textit{OmniInsert} outperforms closed-source commercial solutions. 

%% file: supp_sections/0_implementation.tex
\appendix
\section{Implementation Details} \label{sec: Implementation}

\subsection{Detailed Parameters}
The hyper-parameters used in our experiments are set as follows: 
\begin{itemize}
    \item For the IPO loss, we set the hyper-parameters $\gamma$, $\lambda$ and $\beta$ to 10, 1, 1, respectively, optimizing the model in phase 4. 
    \item For the Subject-Focused Loss (SL), we set $\lambda_1$ and $\lambda_2$ as 1, 1, respectively. 
\end{itemize}

\subsection{Training and Inference Details}
\noindent\textbf{Training time.} The phase-1 subject-to-video training takes about 70k iterations (nearly 2700 A100 GPU hours). The phase-2 pretraining for the MVI task costs about 30k iterations (nearly 1500 A100 GPU hours), and the phase-3 refinement costs about 10k iterations (nearly 500 A100 GPU hours). The last phase-4 preference optimization takes about 8k iterations (nearly 2300 A100 GPU hours).

\noindent\textbf{Training Parameters.} To avoid the computational expense of full fine-tuning, we integrate the LoRA mechanism into the DiT blocks, configuring it with a rank of 256. This results in a total of 600M trainable parameters. 

\noindent\textbf{Training Data Details.} For the training of the phase-1 subject-to-video model and the phase-2 MVI task pretraining, we use a large-scale training dataset containing about 1M samples. Mark data generated by the introduced RealCapture Pipe, SynthGen Pipe (T2I + I2V + subject remove), SynthGen Pipe (video editing), and SimInteract Pipe as \textbf{(a)}, \textbf{(b)}, \textbf{(c)}, \textbf{(d)}, respectively. The actual ratio of different types of data (type \textbf{(a)} : type \textbf{(b)} : type \textbf{(c)} : type \textbf{(d)}) in phase 1 and phase 2 is set to 5:2:2:1. For the training of phase-3 refinement, the dataset contains about 50k samples and the ratio of different data is 3:3:3:1. For the training of phase-4 preference optimization, the training dataset has about 0.5k good-bad paired data, and the ratio is set to 3:3:3:1.

\noindent\textbf{Inference time.} 
Generating a single 5-second 480P video (121 frames) with our proposed model takes approximately 90 seconds using 8 NVIDIA A100 GPUs. Further discussion on potential improvements to inference speed can be found in Sec.~\ref{sec: limitation}.

\subsection{User Study}
To compare with the baseline methods, we conduct a user study as part of the evaluation. The survey randomly presented 40 sets of generated results to each participant. Fig.~\ref{fig: supp_user_study} displays a screenshot from our user study, showcasing a set of generated results. From left to right, it shows the reference subject, the source video, and the results from three competing methods in random order, paired with the corresponding text prompt. We ask each participant four questions:
\begin{enumerate}
    \item \textit{Which result appears to have the highest consistency with reference subjects? }
    \item \textit{Which result best matches the prompt `[prompt]'?}
    \item \textit{Which result appears to have the highest insertion rationality? }
    \item \textit{Which result matches your best choice based on comprehensive considerations? }
\end{enumerate}

For each set of results displayed in the survey, we ensured that their order was randomly shuffled to prevent bias. Responses where all answers had the same selection and responses with completely identical answers were considered invalid. Finally, we obtained a total of 30 valid surveys to evaluate the model.

%% file: supp_sections/1_bench.tex
\begin{figure*}[tp]
    \centering
    \includegraphics[width=0.95\textwidth]{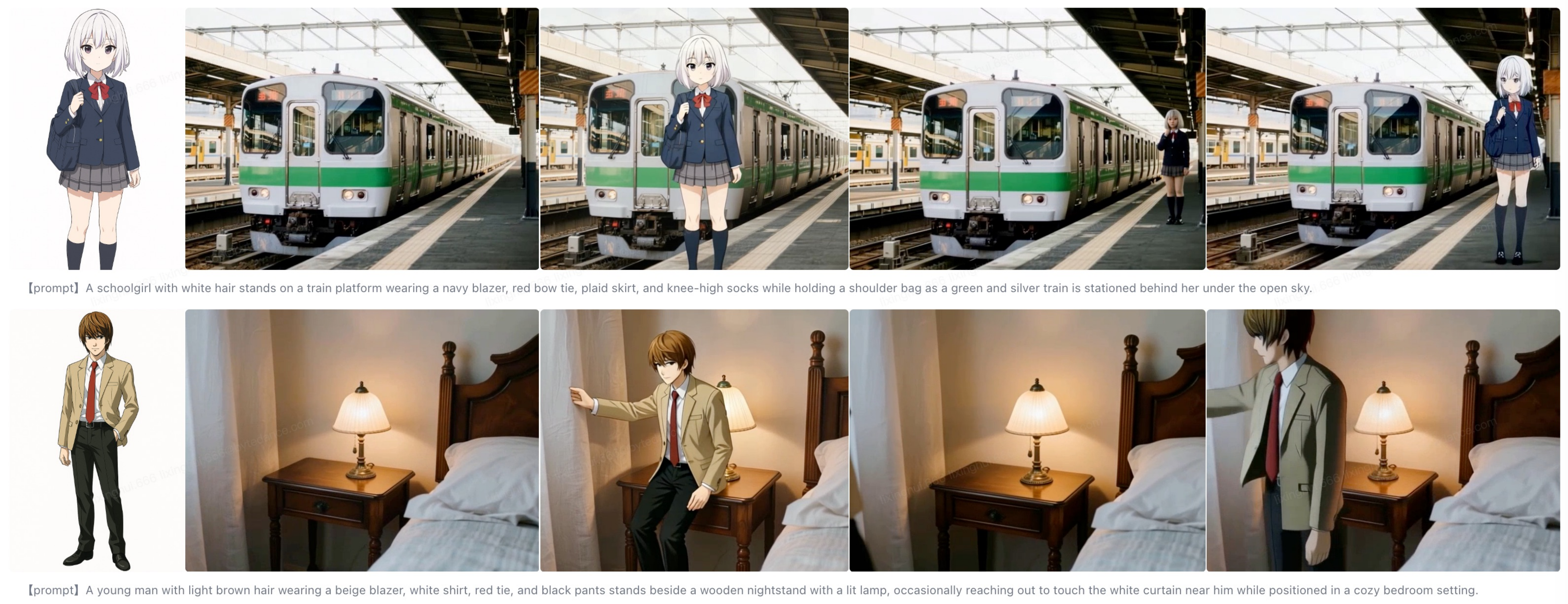}
    \caption{Screenshot of \textbf{user study}.}
    \label{fig: supp_user_study}
\end{figure*}
\begin{figure*}[tp]
    \centering
    \includegraphics[width=0.95\textwidth]{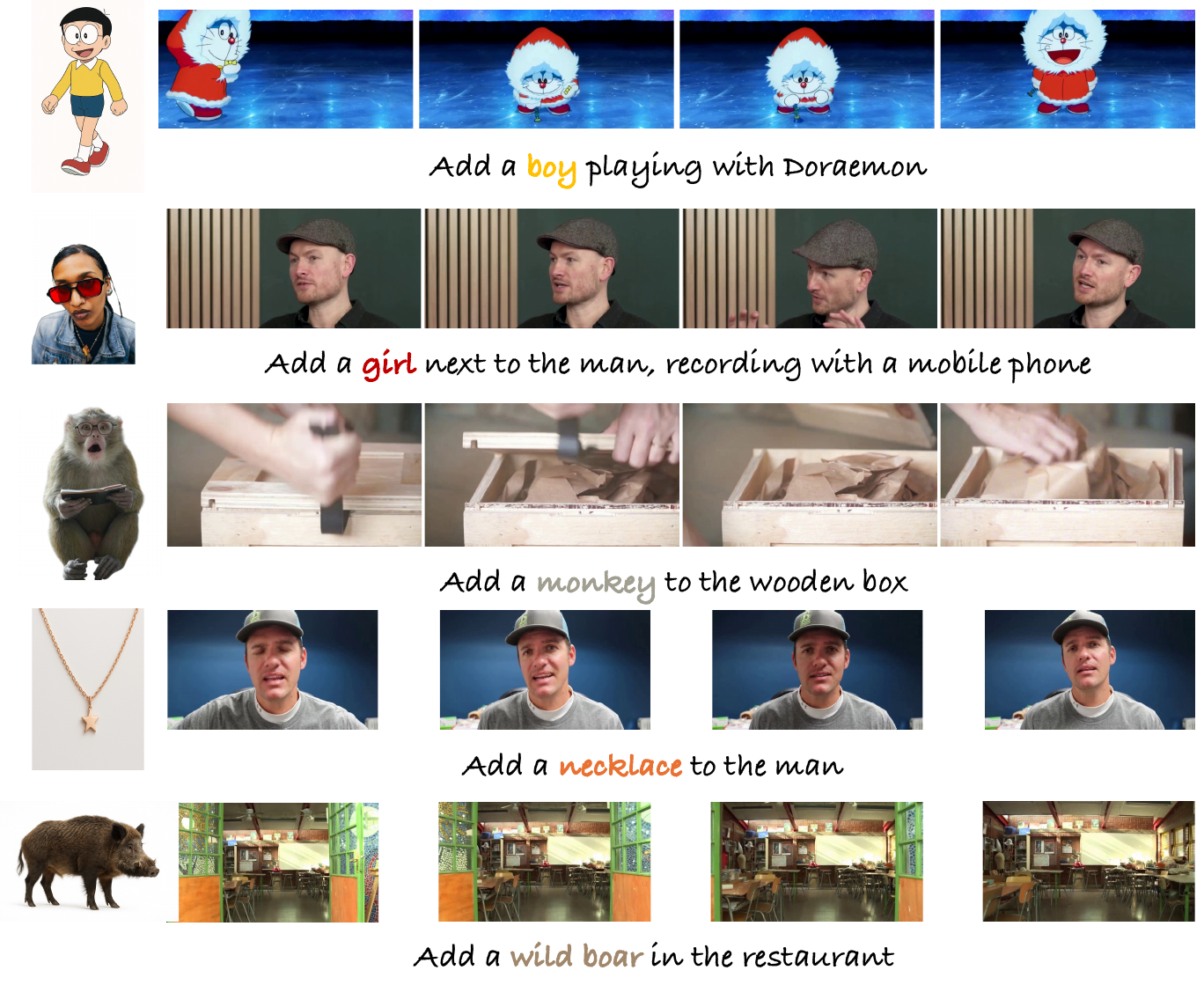}
    \caption{Examples of \textbf{\textit{InsertBench}}.}
    \label{fig: supp_InsertBench}
\end{figure*}
\section{InsertBench Details} \label{sec: InsertBench} 
To address the lack of a benchmark for Mask-free Video Insertion (MVI), we introduce a comprehensive benchmark, \textit{\textbf{InsertBench}}, which consists of 120 videos paired with meticulously selected subjects (suitable for insertion in each video) and the corresponding prompts. Our collected videos span diverse categories, including natural landscapes, indoor environments, transportation scenes, dynamic interactions, animated contexts and wearable scenarios, each comprising 121 frames at 24 fps. As shown in Fig.~\ref{fig: supp_InsertBench}, we carefully select suitable subjects for each video to insert, while considering harmonization and creativity.

%% file: supp_sections/2_more_res.tex
\section{More Visual Results} \label{sec: visual results}
In this section, we provide more visual results of \textit{\textbf{OmniInsert}}. 
Fig.~\ref{fig: supp_res1} and Fig.~\ref{fig: supp_res2} present our inference results, Fig.~\ref{fig: supp_com1} and Fig.~\ref{fig: supp_com2} show comparisons with baselines. Corresponding videos for all results (main paper and supplement) are available in supplementary attachments.
\textbf{We strongly recommend viewing the mentioned dynamic results for a more intuitive understanding of the practical effectiveness of our proposed method.}

All results are generated at 480P resolution based on our proposed benchmark \textit{InsertBench}.
Supplemental results demonstrate enhanced capabilities in diverse aspect ratios (e.g., vertical video) and multi-subject scenarios, further improving our method can achieve stable, semantically coherent insertions with strong prompt fidelity. 
Thanks to our designed data curation pipeline \textit{InsertPipe}, which covers a wide range of scenes and subject categories, the model exhibits excellent robustness across various resolutions and aspect ratios. 
Our designed Condition-Specific Feature Injection (CFI) mechanism injects both video and subject features in a
unified yet efficient manner, ensuring differentiated condition injection. And we integrate the LoRA mechanism into
the DiT blocks, preserving the model’s original prompt following capabilities. 
Furthermore, the Progressive Training (PT) strategy enables the model to effectively achieve subject-scene equilibrium through multi-stage optimization, further enhancing insertion stability. As shown in Fig.~\ref{fig: supp_com1} and Fig.~\ref{fig: supp_com2}, both Pika-Pro~\cite{Pika} and Kling~\cite{Keling} exhibit issues with insertion failures and unnatural insertion artifacts, whereas our method demonstrates superior robustness. 
Additionally, our model is capable of dealing with multi-subject scenarios, a capability that other baseline models either lack or perform poorly. This is achieved through our CFI design, where the IP information of multiple subjects is concatenated along the temporal dimension, enabling a unified training framework with the common single-subject insertion. 

%% file: supp_sections/3_limitations.tex
\section{Limitation and Future Work} \label{sec: limitation}

In this section, we discuss the limitations of our method and potential directions for future work.

\begin{figure}[tp]
    \begin{center}
    \vspace{-0.15in}
    \includegraphics[width=0.8\textwidth]{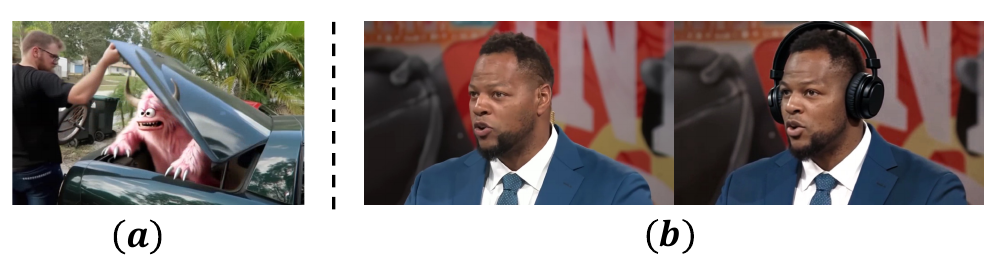}
    \end{center}
    \caption{The failure cases. \textbf{(a)} shows the physically implausible bad cases, such as model interpenetration, and \textbf{(b)} shows the slight color discrepancy between the source video and the inference result.}
    \label{fig: supp_failure}
\end{figure}

\noindent\textbf{Color Fidelity and Physical Plausibility}. 
We observe that our generated results may occasionally exhibit slight color discrepancies compared to the reference video, as well as minor physically implausible phenomena, as illustrated in Fig.~\ref{fig: supp_failure}. Notably, such issues are not unique to our method and are commonly observed across competitive baselines.
Although our designed Insertive Preference Optimization (IPO) in Phase-4 has significantly reduced the challenges, complete resolution is still difficult. 
Due to the dependence of the IPO mechanism on the model's intrinsic high-quality sample generation capability, the pool of selectable high-quality examples may become constrained when such artifacts are prevalent, especially in diversity.
Therefore, incorporating more advanced preference optimization techniques~\cite{azar2024general, Song_Yu_Li_Yu_Huang_Li_Wang_2024} is a crucial direction for future improvement.

\noindent\textbf{Inference Speed.} 
Our method adheres to a standard video diffusion model baseline, with an inference time of approximately 90 seconds for a 480P video with 121 frames. 
In contrast, the inference speeds of the compared bashlines are all above 180s. (Notably, as our compared baselines are all closed-source, we are unable to conduct a fair comparison by running their models offline.) Although our method demonstrates superior inference speed, its performance could be further enhanced by applying general acceleration techniques~\cite{wang2024animatelcm, liang2024looking} for video diffusion models.

\begin{figure*}[tp]
    \centering
    \includegraphics[width=0.85\textwidth]{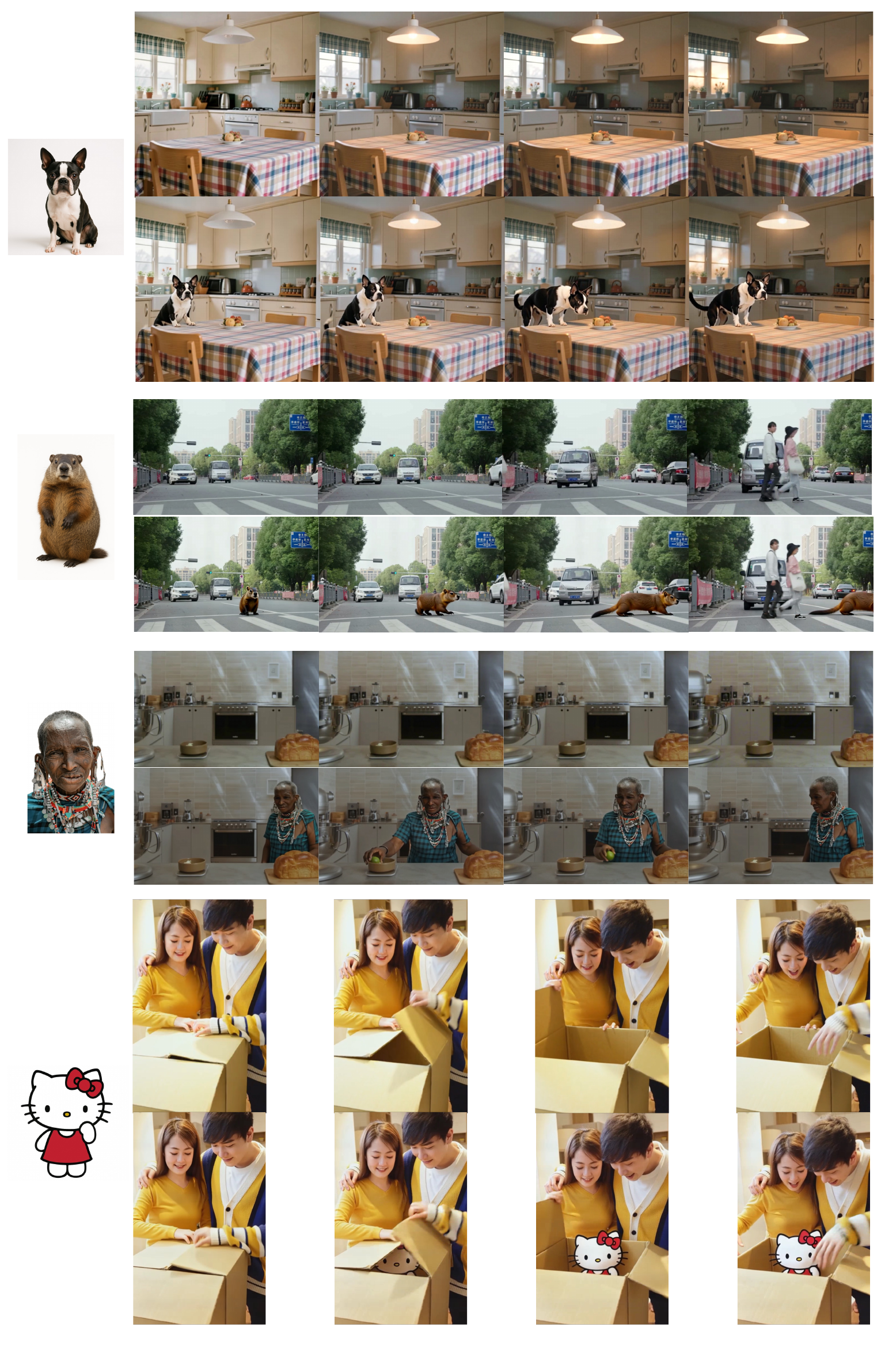}
    \caption{More \textbf{qualitative results I}.}
    \label{fig: supp_res1}
\end{figure*}
\begin{figure*}[tp]
    \centering
    \includegraphics[width=\textwidth]{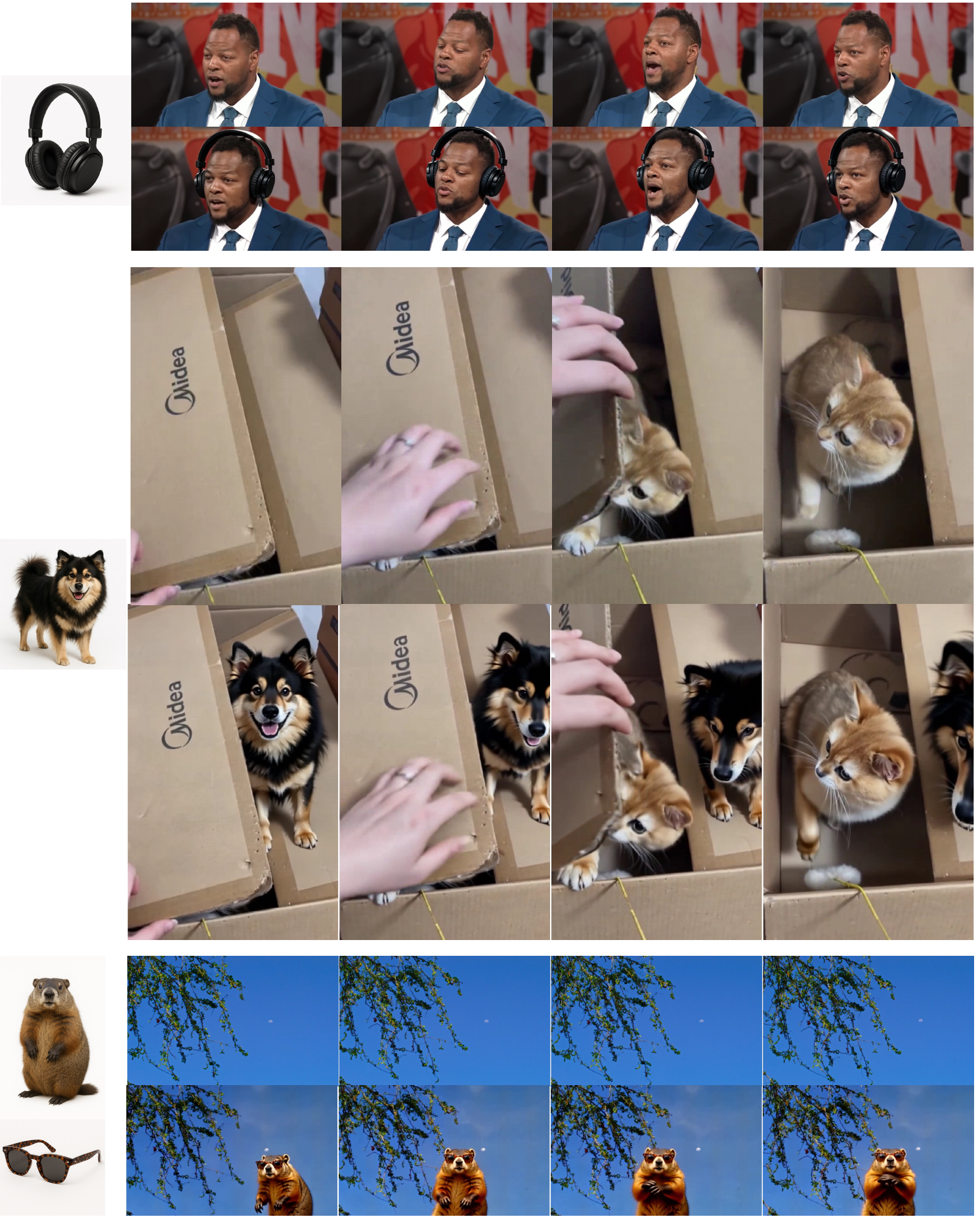}
    \caption{More \textbf{qualitative results II}.}
    \label{fig: supp_res2}
\end{figure*}

\begin{figure*}[tp]
    \centering
    \vspace{-0.15in}
    \includegraphics[width=0.8\textwidth]{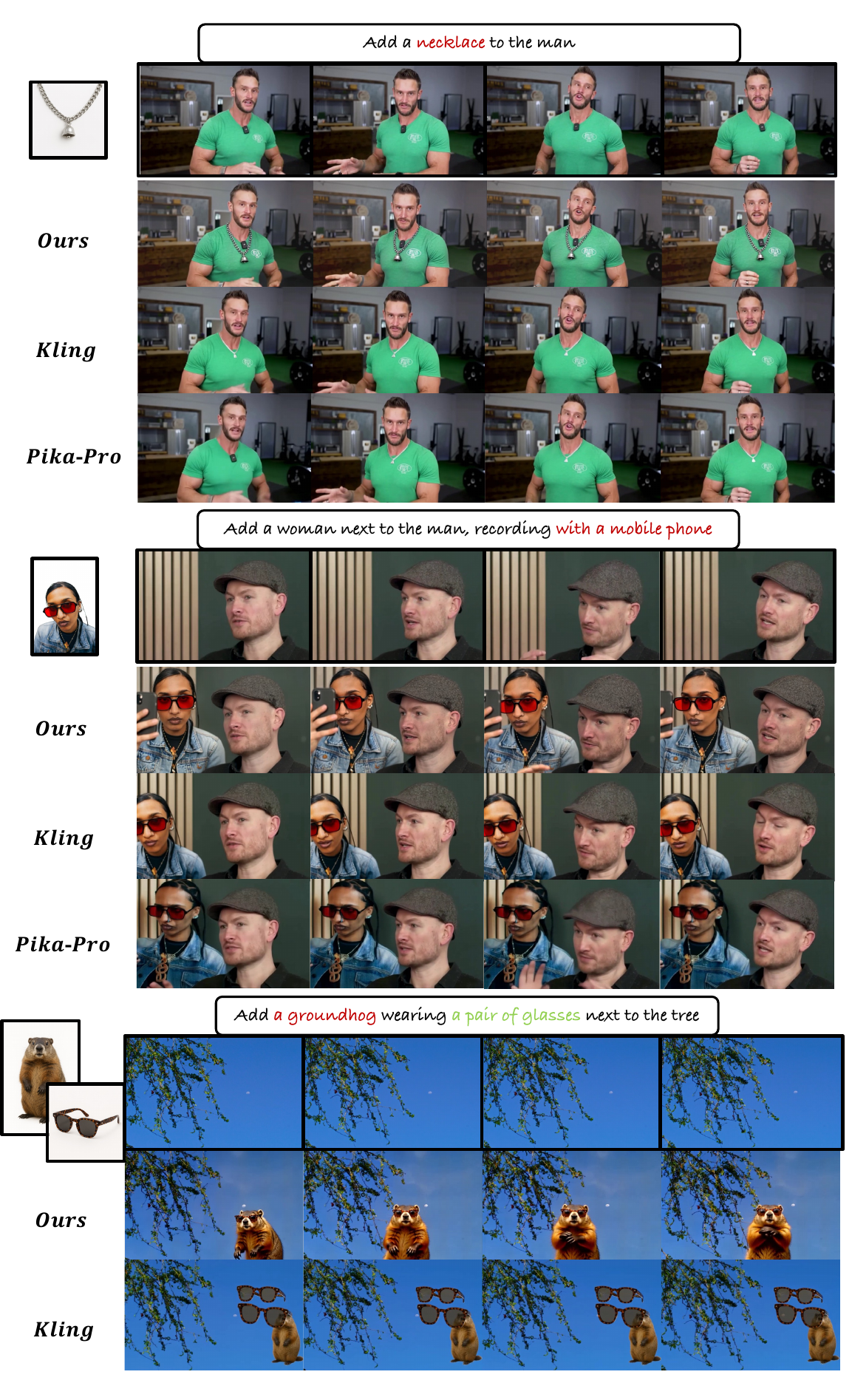}
    \caption{More \textbf{qualitative comparisons I}.}
    \vspace{-2mm}
    \label{fig: supp_com1}
\end{figure*}
\begin{figure*}[tp]
    \centering
    \vspace{-0.15in}
    \includegraphics[width=0.85\textwidth]{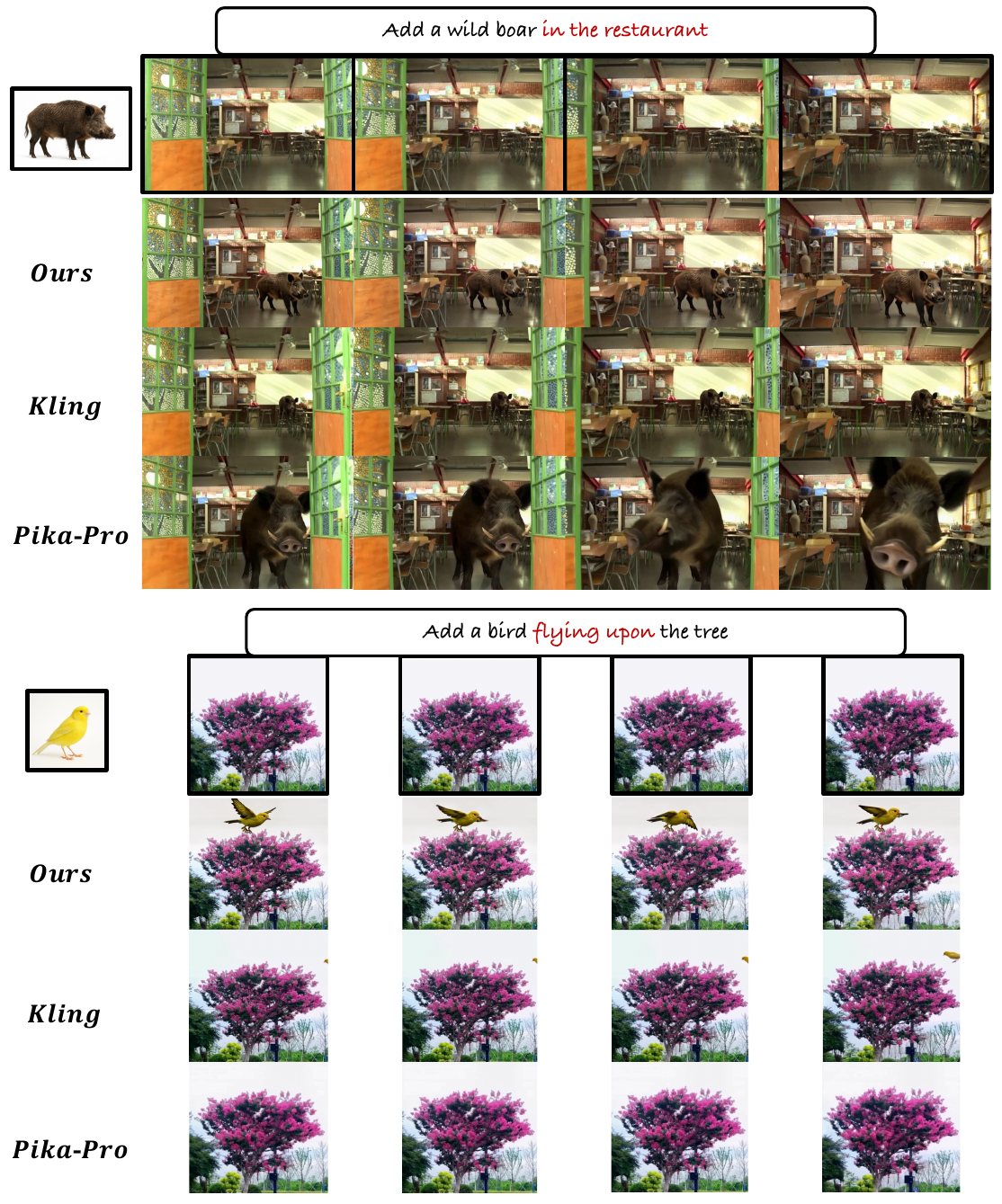}
    \caption{More \textbf{qualitative comparisons II}.}
    \vspace{-2mm}
    \label{fig: supp_com2}
\end{figure*}
